%% file: main.tex
\documentclass{article}

\usepackage{arxiv}

\usepackage[utf8]{inputenc} 

\usepackage{amsfonts}       
\usepackage{nicefrac}       
\usepackage{microtype}      
\usepackage{lipsum}

\usepackage{color}
\usepackage{listings}
\usepackage{booktabs}
\usepackage{array}
\usepackage{graphicx}
\usepackage{longtable}
\usepackage{subfigure}

\usepackage{cite}
\usepackage{url}
\usepackage{fancyhdr}

\usepackage{soul}

\usepackage{float}
\usepackage{listings}
\usepackage{mdwmath}
\usepackage{mdwtab}
\usepackage{multirow}
\usepackage{multicol}
\usepackage{rotating}
\usepackage{setspace}
\usepackage[utf8]{inputenc}
\usepackage{lineno}
\usepackage{listings}

\usepackage{mdwmath}
\usepackage{mdwtab}
\usepackage{multirow}
\usepackage{multicol}
\usepackage{array}
\usepackage{booktabs}

\usepackage{enumitem}
\usepackage{xspace}
\usepackage[export]{adjustbox}
\usepackage{graphicx}
\usepackage{color,soul}
\usepackage{rotating}
\usepackage{setspace}
\usepackage{amsmath} 
\usepackage{amssymb}
\usepackage{float}
\usepackage{xcolor}

\usepackage{hyperref}
\usepackage{url}

\title{Towards Responsible Generative AI:\\ A Reference Architecture for Designing Foundation Model based Agents}

\author{Qinghua Lu, Liming Zhu, Xiwei Xu, Zhenchang Xing, Stefan Harrer, Jon Whittle\\
Data61, CSIRO, Australia\\
\textit{Email: \{firstname\}.\{lastname\}@data61.csiro.au}
}

\begin{document}

\maketitle

\begin{abstract}
Foundation models, such as large language models (LLMs), have been widely recognised as transformative AI technologies due to their capabilities to understand and generate content, including plans with reasoning capabilities. Foundation model based agents derive their autonomy from the capabilities of foundation models, which enable them to autonomously break down a given goal into a set of manageable tasks and orchestrate task execution to meet the goal. Despite the huge efforts put into building foundation model based agents, the architecture design of the agents has not yet been systematically explored. Also, while there are significant benefits of using agents for planning and execution, there are serious considerations regarding responsible AI related software quality attributes, such as security and accountability. Therefore, this paper presents a pattern-oriented reference architecture that serves as guidance when designing foundation model based agents. We evaluate the completeness and utility of the proposed reference architecture by mapping it to the architecture of two real-world agents. 

\end{abstract}

\textbf{Key terms - } Foundation model, large language model, LLM, agent, architecture, pattern, responsible AI, AI safety.

\section{Introduction}
Foundation models (FMs)~\cite{bommasani2021opportunities}, such as large language models (LLMs), have been widely recognised as transformative generative artificial intelligence (GenAI) technologies due to their remarkable capabilities to understand and generate content. FMs are pretrained on massive amounts of data and can be adapted to perform a wide variety of tasks and significantly improve productivity. Significant efforts have been placed on utilising FMs' human-like reasoning capabilities for a diverse range of downstream tasks, such as question answering and information summary. However, it is crucial to acknowledge that FMs exhibit inherent limitations, particularly when facing complex tasks. Users are required to provide prompts at each individual step, which can be inefficient and prone to errors. 
There have been recently a rapidly growing interest in the development of FM-based autonomous agents~\cite{xi2023rise, wang2023survey}, such as Auto-GPT\footnote{https://github.com/Significant-Gravitas/Auto-GPT} and BabyAGI\footnote{https://github.com/yoheinakajima/babyagi}. With autonomous agents, users only need to provide a high-level goal, rather than providing explicit step-by-step instructions. These agents derive their autonomy from the capabilities of FMs, enabling them to autonomously break down the given goal into a set of manageable tasks and orchestrate task execution to fulfill the goal. While huge efforts have been put on building FM-based agents, the architecture design of the agents has not yet been systematically explored. Many reusable solutions have been proposed to address the diverse challenges for designing FM-based agents, which motivates the design of a reference architecture for FM-based agents.

On the other hand, 
there are significant challenges in responsible AI when designing FM-based agents~\cite{lu2023framework, lu2023towards}. First, autonomy is a core feature enabled by FM-based agents. These agents can infer human intentions and goals, either explicitly or implicitly, from multimodal context information, generate plans, use external tools/systems, cooperate with other agents, and may even create new tools and agents. In such cases, the accountability for actions taken by the agents may be shared among the agent owner, the FM provider, and various providers of external tools/agents. 
Second, enabling accountability necessitates underlying supporting mechanisms for traceability. 
Third, the goals or instructions set by humans for the agents, as well as the agents' outputs and behaviour, including interaction with external tools and other agents, must be trustworthy and responsible.  


Therefore, we have performed a systematic literature review (SLR) on FM-based agents. Based on the review results, a collection of architectural components and patterns have been identified to address different challenges of agent design. This paper presents a pattern-oriented reference architecture (RA) that serves as architecture design guidance for designing FM-based agents. We evaluate the completeness and utility of the proposed reference architecture by mapping it to the architecture of two real-world agents: MetaGPT~\cite{hong2023metagpt} and HuggingGPT~\cite{shen2023hugginggpt}. 
The contribution of this paper includes:
\begin{itemize}
    \item A reference architecture for FM-based agents that can be used as a template to guide the architecture design.
    \item A collection of architectural patterns that can be utilised in the architecture design of FM-based agents to ensure trustworthiness and address responsible AI related software quantities.
\end{itemize}


\section{Methodology}
 This section presents the methodology employed in this study. An empirically-grounded design methodology has been
adopted~\cite{galster2011empirically}. 
Our initial step involving determining the type of our reference architecture. We decided to design an industry-crosscutting, classical, facilitation reference architecture. Here, “industry-crosscutting” means that the reference architecture can span multiple industries, ”classical” indicates that its development is based on existing FM-based agents, and “facilitation” signifies that its aim is to guide the future design of FM-based agents, whether within a single organisation or across multiple organisations. Our design strategy is a combination of “research-driven” and “practice-driven”, as the design of this reference architecture is founded mainly on the findings of an SLR which includes both academic research papers and industry papers. Our project experiences in designing the FM-based scientific discovery agent and responsible AI copilot have also been beneficial for this study, particularly in multimodal context engineering and responsible AI. The third step is the empirical acquisition of data. We performed an SLR to identify the relevant studies. The key search terms include foundation model and agent. The supportive terms are large language model, LLM, FM. We finally identified 57 studies after conducting the paper search, snowballing, and quality assessment. 
Based on the acquired data, we constructed a
reference architecture for FM-based agents by integrating the architectural patterns and components identified.
We annotated different patterns that can
lead to the instantiation of various concrete architectures for FM-based agents. In the final step, the evaluation of our proposed reference architecture was carried out by reviewing two real-world agents. We mapped the architectural components of the two agents onto the proposed reference architecture, to evaluate that the reference architecture can be transformed into meaningful concrete architectures.

\begin{figure*}
\centering
\includegraphics[width=0.65\textwidth]{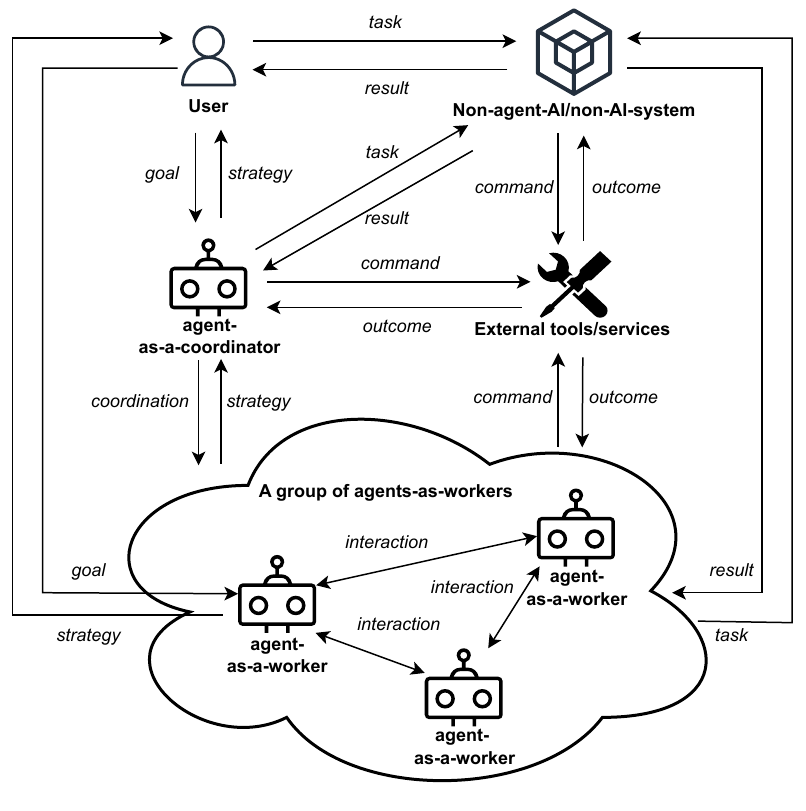}
\caption{Architecture of an agent-based ecosystem.} \label{fig:eco}
\vspace{-2ex}
\end{figure*}

\section{Reference Architecture}
\begin{figure*}
\centering
\includegraphics[width=\textwidth]{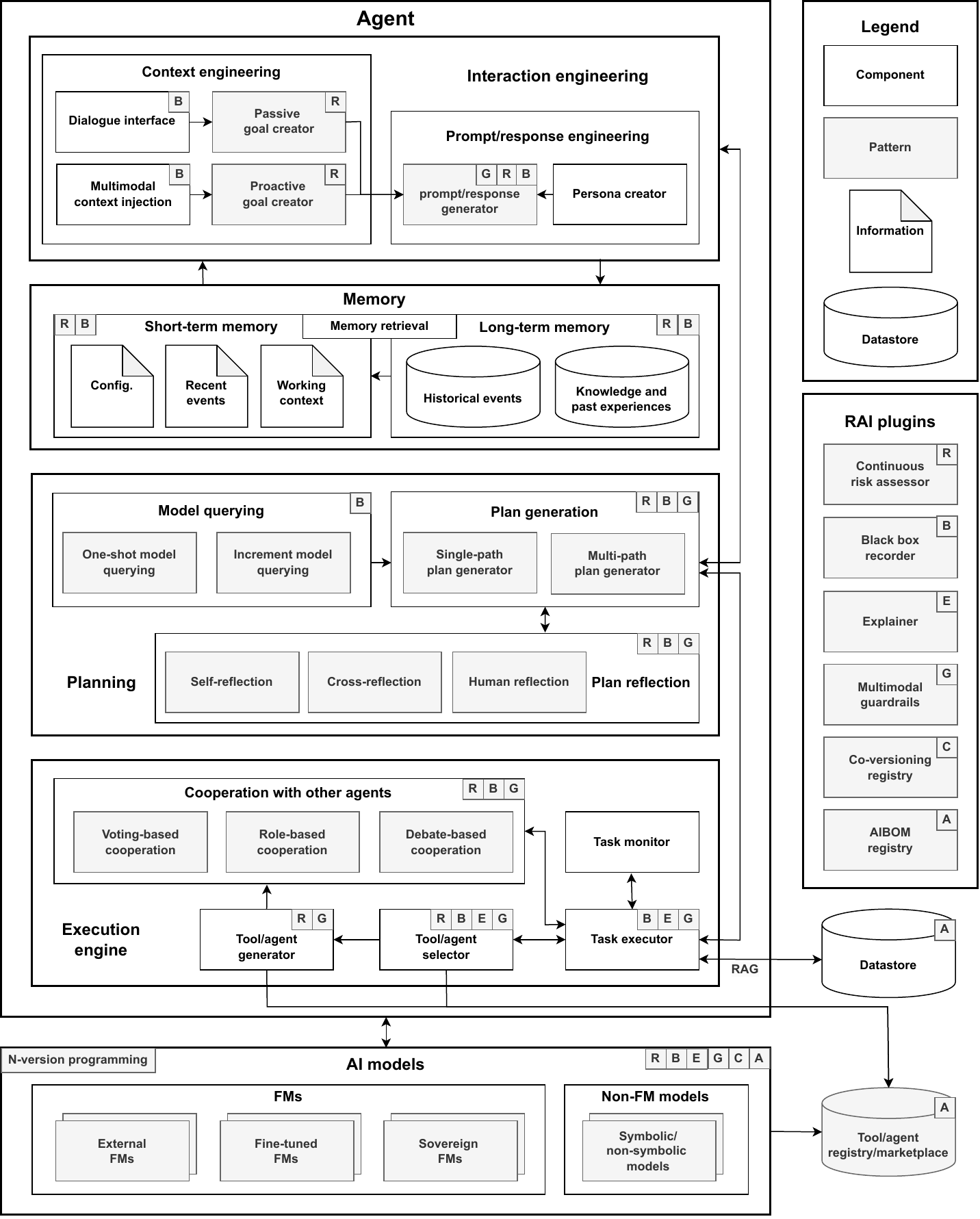}
\caption{Reference architecture for designing an autonomous agent.} \label{fig:agent}
\vspace{-2ex}
\end{figure*}

Fig.~\ref{fig:eco} provides an architectural overview of an agent-based ecosystem. Users define high-level goals for the agents to achieve. The agents can be categorised into two types~\cite{liu2023bolaa, liu2023dynamic, he2023chateda}: agent-as-a-coordinator and agent-as-a-worker. Agents in the coordinator role primarily formulate high-level strategies and orchestrate the execution of tasks by delegating task execution responsibilities to other agents, external tools, or non-agent systems. On the other hand, agents in the worker role need to generate strategies and execute specific tasks in line with those strategies. To complete these tasks, agents in the worker role may need to cooperate or compete with other agents, or call external tools or non-agent AI/non-AI systems. Fig.~\ref{fig:agent} illustrates the reference architecture of an FM-based agent.

\subsection{Interaction engineering}
Interaction engineering comprises two components: context engineering and prompt/response engineering. Context engineering is designed to collect and structure the context in which the agent operates to understand the user's  goals~\cite{xia2023towards}, while prompt/response engineering is responsible for generating prompts/responses, enabling the FM-based agents to successfully achieve the human's goals. 
There are two patterns that can be applied for comprehending and shaping the goals: passive goal creator and proactive goal creator. Passive goal creator analyses the user's articulated goals, as described through text prompts submitted by the user via the dialogue interface~\cite{liu2023ai, kannan2023smart}. Conversely, proactive goal creator goes beyond the explicit user text prompt and anticipates the user's goals by understanding the user interface (UI) of relevant tools and human interaction~\cite{zeng2023gesturegpt}. This is facilitated through the analysis of multimodal context information, including screen recording~\cite{zhao2023seehow}, mouse clicks\footnote{https://github.com/ddupont808/GPT-4V-Act}\footnote{ https://www.loom.com/share/9458bcbf79784162aa62ffb8dd66201b}, typing, eye tracking, gestures~\cite{zeng2023gesturegpt}, document annotations and notes. In the multi-agent environment, proactive goal creator can also anticipate other agents' decisions and formulate plans proactively~\cite{zhang2023proagent}. Both passive goal creator and proactive goal creator need to consider the personas created by the users or the other agents through the persona creator~\cite{wang2023unleashing, li2023camel}. The persona can include the roles, communication style, expertise capabilities, limitation boundaries, etc.
Once the goal is confirmed by the user, prompt/response generator automatically generates the prompts/responses with constraints and specifications, defining the desired input or output content and format in alignment with the the ultimate goal. A prompt or response template is often used in the prompt/response generator as a factory that creates prompt or response instances from the template~\cite{zhao2023expel, schumann2023velma}. The template provides a structured way to standardise the queries and responses, which can improve the response accuracy and interoperability with external tools or agents.

\subsection{Memory}
The agent's memory stores current context information and historical data and knowledge that have been accumulated over time to inform planning and actions. The memory is structured using short-term memory and long-term memory. 
Short-term memory refers to the information within the context window of the FM, which is in-context and can be accessed by the FM during inference. The information stored in short-term memory includes configurations, recent events, and working context~\cite{packer2023memgpt, jin2023surrealdriver, zhao2023expel, zhang2023building}. The configurations encompass the agent's persona, capabilities, constraints, function description, etc. The events cover user interactions (e.g., prompts, clicks), system messages, and other events that have occurred recently. The working context refers to the status of the task or set of tasks that the agent is currently working on.
The length of short-term memory is limited by the context window length of the FM, which is the length of text the FM can take as input when generating a response. For example, the latest version of GPT-4 has a maximum limit of 128k tokens, equivalent to approximately 300 pages of text\footnote{https://openai.com/blog/new-models-and-developer-products-announced-at-devday}.  It is challenging to provide prompts containing data larger than the token limit.  
To enhance the storage capacity of memory, long-term memory refers to the information maintained outside the context window of the FM. This long-term memory stores two types of information~\cite{packer2023memgpt, jin2023surrealdriver}: the entire history of events processed, and the knowledge and past experiences (e.g., observations, thoughts, behaviour). For the FM to process this information, it must be explicitly selected from long-term memory through memory retrieval and moved into short-term memory.


\subsection{Planning}
To achieve the user's goal, the agent needs to work out strategies and make a plan accordingly~\cite{ye2023proagent}. There are two design patterns for plan generation: single-path plan generator and multi-path plan generator. The single-path plan generator orchestrates the generation of intermediate steps leading to the achievement of the user's goal. Each step is designed to have only one subsequent step, such as Chain-of-Thought~\cite{wei2022chain}. Self-consistency asks the FM several times and selects the most consistent answer as the final answer~\cite{wang2023rcagent}. On the other hand, multi-path plan generator allows for the creation of multiple choices at each step. Each intermediate step may lead to multiple subsequent steps. Self-consistent Chain-of-Thought~\cite{wang2022self} and Tree of Thought~\cite{yao2023tree} exemplify this design pattern. 
Both the single-path plan generator and the multi-path plan generator can involve two design patterns of model querying: one-shot model querying and incremental model querying. In one-shot model querying, such as Chain-of-Thought~\cite{wei2022chain} and self-consistent Chain-of-Thought~\cite{wang2022self}, the FM is accessed in a single instance to efficiently generate all the necessary reasoning steps for the plan. The incremental model querying, such as Tree of Thought~\cite{yao2023tree}, involves accessing the FM at each step of the plan generation process, which enables a more iterative and interactive planning process that can adapt to evolving circumstances or changing user requirements.
Plan reflection allows the agent to incorporate feedback to refine the plan~\cite{park2023generative} through three design patterns: self-reflection, cross-reflection, and human reflection. Self-reflection enables the agent to generate feedback on the plan and provide refinement guidance from themselves~\cite{yao2022react, madaan2023self, sumers2023cognitive, shinn2023reflexion}. Cross-reflection uses different agents or FMs to provide feedback and refinement on the plan~\cite{chen2023interact, shinn2023reflexion, talebirad2023multi}. The agent can collect feedback from humans to refine the plan, which can effectively make the plan aligned with the humans preference~\cite{huang2022inner, sarch2023open}.

\subsection{Execution engine}
Once the plan is determined, the role of the execution engine is to put the plan into action. The task executor is responsible for performing the tasks outlined in the plan. A task monitor is necessary to monitor the task's execution status and manage the tasks queued for execution~\cite{chen2023autoagents, nascimento2023self}.
Apart from directly utilizing internal knowledge of FMs or extracting external knowledge through retrieval augmented generation (RAG) to guide the action execution, the task executor can cooperate with other agents or leverage external tools~\cite{ruan2023tptu, kong2023tptu}, which expands the capabilities of the agent. The tool/agent selector~\cite{xie2023openagents, talebirad2023multi} can perform a search in the tool/agent registry/marketplace~\cite{ruan2023tptu, wang2023voyager,kong2023tptu} or on the web to find the relevant tools and agents to complete the tasks. Humans can trade or hire specialised agents in the marketplace. The owners of these agents can receive incentives as they build agents' memory based on their own knowledge or skills. An FM-based ranker can be applied to analyse the performance of the tools/agents and identify the best ones~\cite{liu2023dynamic}. The tool/agent generator can automatically create tools and agents based on natural language requirements~\cite{qian2023creator, chen2021evaluating}. 

There are three design patterns to support multi-agent cooperation: voting-based cooperation, role-based cooperation, and debate-based cooperation. In the voting-based cooperation, agents can freely provide their opinions and reach consensus through voting~\cite{hamilton2023blind}. The cooperation can be also reached by assigning different roles to agents through role-based cooperation~\cite{li2023camel, qian2023communicative, li2023metaagents, qian2023communicative, li2023camel}. One special type of role-based cooperation is the competitive auction mode. This involves two roles: the auctioneer who proposes a task and orchestrates the auction process, and the bidders who actively participate in the bidding~\cite{chen2023put, zhao2023competeai}.
Debate-based cooperation integrates concepts from game theory, where one agent can receive feedback from other agents~\cite{du2023improving, liang2023encouraging} and adjust the thoughts and behaviours until consensus is reached~\cite{chen2023multi, tang2023medagents}.


\subsection{Responsible AI (RAI) plugins}
To ensure responsible AI, a set of patterns can be adopted as plugins~\cite{lu2022responsible}. 
A continuous risk assessor~\cite{lee2023qb4aira,xia2023principles} continuously monitors and assesses AI risk metrics to prevent the misuse of the agent and to ensure the trustworthiness of the agent. 
A black box recorder~\cite{lu2023framework} records the runtime data, which can be then shared with relevant stakeholders to enable transparency and accountability. The recorded data includes the input, output, and intermediate data for each component within the architecture, such as the input and output of the FMs, external tools, or other RAI plugins (e.g. guardrails). 
All these data need to be kept as evidence with the timestamp and location data, e.g., using a blockchain-based immutable log.

A specific type of monitoring is called guardrails, which is a layer in between FMs or fine-tuned FMs and other components or systems. They are designed to control the inputs and outputs of FMs to meet specific requirements, such as user requirements, ethical standards, laws. Guardrails can be built on an RAI knowledge base, or RAI narrow models or RAI FMs. The RAI FMs can be fine-tuned or can call upon an RAI knowledge base to support RAI controls. There can be different types of guardrails. Input guardrails are applied to the inputs received from users, such as refusing or modifying user prompts. For example some users' prompts may contain personally identifiable information (PII) and need to be removed by employing data de-identification and anonymisation~\cite{abbasian2023conversational} before being sent to the FMs. Output guardrails focus on the output generated by the FM, such as modifying the output of the FM or preventing certain outputs from being returned to the user. RAG guardrails are used to ensure the retrieved data is appropriate, either by refusing or modifying the retrieved data. Execution guardrails are applied to input/output of the narrow AI models or external tools that the FM may invoke for action execution. Whitelists and blacklists can be established to identify actions that are permitted and prohibited respectively~\cite{yang2023plug}. During the workflow execution, intermediate guardrails can be used to guarantee that each intermediate step meets the necessary criteria. This involves checking if an action should be carried out or determining if the FM should be invoked, or deciding if a predefined response should be used instead, etc.

To provide comprehensive monitoring and control, multimodal guardrails can be designed to prevent inappropriate multimodal inputs sent to the FM, whether those inputs are from the users or other software components or external tools or models. 
Prevent inappropriate multimodal outputs generated by the FM itself, whether those outputs are sent to the user or other software components or external tools or models.
For example, if UIs are generated dynamically on demand, multimodal guardrails can effectively flag any UI elements that fail to meet specific requirements such as standards.


The explainer's role is to articulate the agent's roles, capabilities, limitations, the rationale behind its intermediate or final outputs~\cite{wang2023describe}, and ethical or legal implications~\cite{schwartz2023enhancing}. However, verifying the explanations can be challenging.
The external systems, including tools, agents, FMs, can be associated with an AIBOM that records their supply chain details, including AI risk metrics or verifiable responsible AI credentials~\cite{xia2023trust, xia2023empirical}. The procurement information can be maintained in an AIBOM registry. The agent can refuse to run the external tools, agents or FMs if there is questionable provenance.
As there will be more rapid releases about fine-tuned FMs, there is a trend that multiple FM variations co-exist and are serviced at the same time. The co-versioning registry can be applied to co-version the AI components, such as FMs and fine-tuned FMs~\cite{lu2023towards}.

\subsection{AI models}
When designing the architecture of FM-based applications, including agents, one of the most critical decisions is choosing whether to use an external FM, or a fine-tuned FM~\cite{lin2023swiftsage, kong2023tptu}, or build a sovereign FM in-house from scratch~\cite{lu2023framework, lu2023towards}. 
Using an external FM can be cost-effective and can potentially result in higher accuracy and generalisability to a wider range of tasks. To better perform domain-specific tasks, the parameters of an external FM can be fine-tuned using domain-specific data. Training a sovereign FM in-house from scratch provides complete control over the model pipeline to ensure security and help keep competitive advantage in some domains. However, this requires significant investments in terms of cost and resources. N-version programming~\cite{lin2023swiftsage, nafreen2020architecture, lu2022responsible} can improve the reliability of the agent by using different versions of FMs as basis for its reasoning.

\section{Evaluation}
We evaluate the completeness and utility of the proposed reference architecture by mapping it to two real-world agents: MetaGPT\cite{hong2023metagpt} and HuggingGPT~\cite{shen2023hugginggpt}. MetaGPT enables multi-agent collaborations to streamline the software engineering workflow. HuggingGPT leverages FMs to connect different AI models based on their function descriptions found in the Hugging Face machine learning community to address various tasks. 
The interactions between humans and MetaGPT are sent through the dialogue interface. The persona creator is realised by creating specialised roles based on the Role class. 
The passive goal creator is implemented by incorporating the goal into the specialised role. 
The prompt generator formulates a structured prompt template for each action conforming to role-specific standards.
Both short-term memory and long-term memory are integrated into the design. Each agent is created using role-specific prompts to establish a role context. An individual memory cache is maintained by each agent to index subscribed messages by their content, sender, and recipient, while a shared memory pool is designed for the environment.   
Single-path plan generation is applied through increment model querying. Complex tasks are broken down into smaller, manageable sub-tasks for individual agents to address. Key information is extracted from the environment, stored in memory, and employed to inform reasoning and subsequent actions. 
A feedback mechanism continuously refines the code using its own memory on historical execution and debugging. 
Diverse roles are assigned to various agents and establish effective role-based cooperation. Task executor uses role-specific interests to gather relevant information and executes its action once it has received all the necessary prerequisites. 
Different roles are allocated by tool/agent generator to various agents with unique skills, each contributing specialised outputs for particular tasks.  
Guardrails define the operational boundaries or constraints that the role must follow during the execution of actions. 
The explainer displays the task-related thoughts and the resulting artifacts on the interface. 
The external FM, GPT4-32k, serves as the underlying FM.


\begin{table*}[tbp]
\footnotesize
\centering
\caption{Mapping components in the architecture of MetaGPT and HuggingGPT.}
\label{tab:comparison}
\begin{tabular}{p{0.15\columnwidth}p{0.425\columnwidth}p{0.375\columnwidth}}
\toprule

{\bf Component} &
\multicolumn{1}{c}{\bf  MetaGPT} &
\multicolumn{1}{c}{\bf  HuggingGPT} \\
\midrule

\multirow{1}{0.14\columnwidth}{Persona creator} & Specialised roles can be created from the fundamental Role class. & N/A\\
\cmidrule(l){1-3}

\multirow{1}{0.14\columnwidth}{Dialogue interface} &  Receiving inputs sent by humans and responds to it. & Generating responses to address user requests.\\
\cmidrule(l){1-3}

\multirow{2}{0.14\columnwidth}{Passive goal creator} & The goal included in the specialised role represents the main objective that the role aims to achieve. & User requests include complex intents that can be interpreted as their intended goals. \\
\cmidrule(l){1-3}

\multirow{1}{0.14\columnwidth}{Prompt generator} & Each action is provided with a prompt template that conforms to the standards for the role. & A task template is provided to guide the FM to analyse user requests and parse tasks accordingly through field filling.\\
\cmidrule(l){1-3}

\multirow{2}{0.14\columnwidth}{Short-term memory} & Each role is created using specialised role prompts to establish a role context. & Model descriptions are integrated into prompts, allowing the FM to select the appropriate models for task execution.\\
\cmidrule(l){1-3}

\multirow{2}{0.14\columnwidth}{Long-term memory} & An individual memory cache is maintained by each role, while a shared memory pool is designed for the environment.   & N/A\\
\cmidrule(l){1-3}

\multirow{2}{0.14\columnwidth}{Single-path plan generation} & Complex tasks are broken into smaller, manageable sub-tasks for individual agents to complete. & User requests are decomposed into a set of structured tasks with dependencies and execution orders.\\
\cmidrule(l){1-3}

\multirow{2}{0.14\columnwidth}{Increment model querying} & Significant information is extracted from the environment, stored in memory, and used to guide reasoning and subsequent actions.  & Generating results by recursively querying the FM.\\
\cmidrule(l){1-3}

\multirow{1}{0.14\columnwidth}{Self-reflection} & A feedback mechanism debugs and executes code at runtime.
& N/A\\
\cmidrule(l){1-3}

\multirow{2}{0.14\columnwidth}{Role-based cooperation} & Establishing collaboration among multiple roles.  & N/A\\ \\
\cmidrule(l){1-3}

\multirow{1}{0.14\columnwidth}{Task executor} & Each agent extracts relevant information and executes its action once it has received all the necessary prerequisites. & Running each selected model and delivering the results.\\
\cmidrule(l){1-3}

\multirow{1}{0.14\columnwidth}{Tool/agent generator} & Allocating different roles to various agents with unique skills.  & N/A\\
\cmidrule(l){1-3}

\multirow{1}{0.14\columnwidth}{Tool/agent detector} & N/A & Identifying relevant models based on their description.\\
\cmidrule(l){1-3}

\multirow{1}{0.14\columnwidth}{Guardrails} & The constraints specify limitations or rules the role must follow during the execution of actions. & N/A\\
\cmidrule(l){1-3}

\multirow{1}{0.14\columnwidth}{Black box recorder} & N/A & The workflow logs and chat history are maintained.\\
\cmidrule(l){1-3}

\multirow{1}{0.14\columnwidth}{Explainer} & The interface displays task-specific thoughts and output artifacts. & A summary of logs is generated for the user.\\
\cmidrule(l){1-3}

\multirow{1}{0.14\columnwidth}{External FM} & GPT4-32k. & GPT-3.5-turbo, text-davinci-003 and GPT-4.\\

\bottomrule
\end{tabular}
\end{table*}

HuggingGPT actively generates responses that address user requests through the dialogue interface.
The passive goal creator is implemented by interpreting user's intents as clearly defined goals. 
The prompt generator creates a task template to direct the FM in processing user requests and structuring tasks via field filling.
HuggingGPT only supports short-term memory. Model descriptions are incorporated into prompts, enabling the FM to choose the appropriate models for solving tasks.
Single-path plan generation coupled with increment model querying is employed for planning. User requests are analysed and decomposed into an ordered sequence of tasks, each with its own prerequisites. The steps are determined by recursively querying the FM.
Tool/agent detector identifies the relevant AI models using the model descriptions. Task executor runs each selected model and produces the results.
The black box recorder maintains the workflow logs, while the explainer synthesises the logs into a comprehensive summary for the user.
HuggingGPT leverages external FMs, including GPT-3.5-turbo, text-davinci-003 and GPT-4, which serve as primary models in its experiments.

Our reference architecture is complete and usable, as two agent architectures can be mapped on the proposed reference architecture. The application of different pattern-oriented components leads to the variability of organisation-specific architectures. 
We observed that the fundamental components in an agent's architecture include prompt engineering, memory, planning, execution engine, and RAI plugins. There are different design options for the components within those fundamental components. For example, MetaGPT employs both short-term and long-term memory, while HuggingGPT only adopts short-term memory.
MetaGPT involves the design with multiple agents. Thus, the interaction with other agent components is needed in the execution engine to support cooperation. Apart from the role-based cooperation pattern employed by MetaGPT, there are two other options: voting-based cooperation and debate-based cooperation.
Both MetaGPT and HuggingGPT have incorporated responsible AI patterns, including guardrails, black box recorder, explainer. For context management, neither employs a proactive goal creator that uses multimodal context information.

\section{Related Work}
The launch of OpenAI's ChatGPT~\cite{openai2023gpt4} in November 2022 set a significant milestone for GenAI, gaining over 100 million users within two months of its release. This triggered an arms race among big tech companies to develop FM-based GenAI products. Google responded with its own GenAI product, Bard\footnote{https://bard.google.com/chat}. By February 2023, Microsoft had already integrated GPT-4 into its search engine, Bing. 
One notable type of FM-based systems is agents. Recently, many researchers have utilised FM as the foundation to build AI agents\cite{he2023chateda, hong2023metagpt}. 
Some studies present the architecture of their agent~\cite{packer2023memgpt, costantino4285483human}. However, these architectures often focuses only on certain components. For example, Packer et al.~\cite{packer2023memgpt} mainly concentrate on the memory design of the agent. There's a lack of a holistic view in architecture design, making it difficult for practitioners to design their own agents. 
Two recent survey papers have performed a comprehensive survey on FM-based agents and presented frameworks for designing them~\cite{xi2023rise, wang2023survey}. However, both papers do not present the methodology, which might result in important studies being missed. Also, according to Bass's definition~\cite{bass2003software}, software architecture comprises software elements, relations among them, and properties of both. These frameworks only list the high-level components supporting their functionality.  There is a lack of system-level thinking, with no explicit identification of software components, relationships among them, and their properties. A reference architecture is needed for practitioners to use as a template and adapt it to design their own version of agents.

\section{Conclusion}
In this paper, we present a pattern-oriented reference architecture which functions as an architecture design template and guides the design of FM-based agents. 
We evaluate the correctness and utility of our proposed reference architecture by mapping it to the architecture of two existing real-world agents. 
We plan to develop decision models for the selection of patterns to further assist the design of FM-based agents.

\input{main.bbl}


\end{document}